\documentclass[runningheads]{llncs}
\usepackage[T1]{fontenc}
\usepackage{graphicx,verbatim}
\usepackage{amsmath, amssymb}
\usepackage{multirow}

\usepackage{hyperref}
\usepackage{color}

\begin{document}
\title{VolDiT: Controllable Volumetric Medical Image Synthesis with Diffusion Transformers}
\titlerunning{VolDiT: 3D Diffusion Transformers}
%
  \author{
  Marvin Seyfarth\inst{1,2,3} \and
  Salman Ul Hassan Dar\inst{1,2,3} \and
  Yannik Frisch\inst{1,2,3} \and
  Philipp Wild\inst{4} \and
  Norbert Frey\inst{2,3} \and
  Florian André\inst{2,3} \and
  Sandy Engelhardt\inst{1,2,3}
  }

  \institute{
  Institute for Artificial Intelligence in Cardiovascular Medicine, 
  Medical Faculty of Heidelberg University, Heidelberg University, Heidelberg, Germany \and
  Department of Cardiology, Angiology, Pneumology, 
  Heidelberg University Hospital, Heidelberg, Germany \and
  DZHK (German Centre for Cardiovascular Research), 
  Partner Site Heidelberg/Mannheim, Heidelberg, Germany \and
  University Medical Center of the Johannes Gutenberg-University Mainz, Mainz, Germany \\
  \email{Marvin.Seyfarth@med.uni-heidelberg.de}
  }
 
 \authorrunning{M. Seyfarth et al.}
  
\maketitle              
\begin{abstract}
Diffusion models have become a leading approach for high-fidelity medical image synthesis. However, most existing methods for 3D medical image generation rely on convolutional U-Net backbones within latent diffusion frameworks. While effective, these architectures impose strong locality biases and limited receptive fields, which may constrain scalability, global context integration, and flexible conditioning. In this work, we introduce \textbf{VolDiT}, the first purely transformer-based 3D Diffusion Transformer for volumetric medical image synthesis. Our approach extends diffusion transformers to native 3D data through volumetric patch embeddings and global self-attention operating directly over 3D tokens. To enable structured control, we propose a timestep-gated control adapter that maps segmentation masks into learnable control tokens that modulate transformer layers during denoising. This token-level conditioning mechanism allows precise spatial guidance while preserving the modeling advantages of transformer architectures. We evaluate our model on high-resolution 3D medical image synthesis tasks and compare it to state-of-the-art 3D latent diffusion models based on U-Nets. Results demonstrate improved global coherence, superior generative fidelity, and enhanced controllability. Our findings suggest that fully transformer-based diffusion models provide a flexible foundation for volumetric medical image synthesis. The code and models trained on public data are available at \url{https://github.com/Cardio-AI/voldit}.


\end{abstract}

\section{Introduction}

Diffusion models \cite{ho2020denoising,rombach2022high} have become the leading approach for high-fidelity image synthesis, often relying on convolutional U-Net backbones \cite{ronneberger2015u} that interleave residual blocks with attention layers. In medical imaging, such models have shown promise for modality translation \cite{kim2024adaptive,li2023ddmm}, data augmentation \cite{khader2023denoising,mao2025medsegfactory,guo2025maisi,wang20253d}, reconstruction \cite{wang20253d}, and segmentation-conditioned synthesis \cite{konz2024anatomically,guo2025maisi}. However, nearly all 3D medical diffusion models still rely on convolutional architectures. While effective, 3D U-Nets impose strong locality biases that may limit global anatomical modeling. Volumetric data requires a consistent structure across slices, yet large receptive fields are costly, and repeated down-/upsampling can attenuate fine-grained details. Vision transformers \cite{dosovitskiy2020image}, in contrast, use global self-attention to capture long-range dependencies at every stage and scale effectively in generative tasks. Diffusion Transformers (DiTs), which replace U-Net backbones with purely transformer-based models, demonstrate strong correlations between model capacity and sample quality \cite{peebles2023scalable,pan2024synthetic}, raising the question: \textit{can a fully transformer-based backbone replace 3D U-Nets for volumetric medical synthesis?} Prior work has explored hybrid transformer-convolutional designs \cite{wang20253d}, where transformers operate locally within sub-volumes. While effective, global volumetric modeling remains dominated by convolution, leaving the potential of fully transformer-based 3D diffusion largely unexplored. Beyond scalability, transformer architectures align naturally with emerging foundation-model paradigms in medical AI, enabling unified token-based representations and flexible conditioning on masks \cite{konz2024anatomically,guo2025maisi}, anatomical priors \cite{kebaili2025multi}, multimodal inputs \cite{kim2024adaptive,li2023ddmm,pan2024synthetic}, or clinical variables \cite{pinaya2022brain}.  Here, we introduce \textbf{VolDiT}, the first purely 3D Diffusion Transformer for volumetric medical image synthesis. Unlike hybrid approaches, VolDiT operates on the entire latent volume, replacing the 3D U-Net backbone with transformer blocks. Using a 3D patchification strategy, we tokenize the latent volume to enable global self-attention across all spatial locations. We further propose a timestep-gated control adapter \textit{TGCA} \cite{t2iadapter}, which maps spatial conditioning inputs (e.g., segmentation masks) into learnable tokens that modulate the transformer layers. This enables precise, token-level control over generated volumes while maintaining scalability. Our contributions are threefold:  
(1) formulation of volumetric medical image synthesis using a fully transformer-based diffusion backbone operating on 3D latent volumes,  
(2) introduction of a timestep-gated control adapter (TGCA) for stable and anatomically precise spatial conditioning, compatible with pretrained transformer backbones.  
(3) comprehensive evaluation including comparisons to U-Net-based latent diffusion and non-diffusion generative models, as well as ablation studies to assess fidelity, diversity, and scalability on medical datasets.

\section{Methodology}

This work serves as the methodological foundation for our concurrently submitted 4D spatiotemporal extension (see Supp. 1), in which the proposed architecture is generalized to model dynamic cardiac MR sequences. In contrast, the present paper focuses exclusively on native 3D volumetric medical image synthesis and introduces VolDiT, the first purely transformer-based 3D Diffusion Transformer operating directly on volumetric tokens. We replace conventional U-Net backbones used in 3D latent diffusion models with global self-attention over 3D patch embeddings. Furthermore, we apply a timestep-gated control adapter that injects segmentation-derived control tokens into the denoising transformer, allowing structured and spatially precise conditional generation. The present study establishes the core 3D transformer diffusion framework and systematically evaluates its advantages over convolutional latent diffusion baselines in both unconditional and conditional volumetric synthesis settings.
\subsection{Latent Volumetric Representation via VQ-GAN}
We adopt a latent diffusion strategy and first learn a compact volumetric representation using a 3D Vector-Quantized GAN (VQ-GAN) \cite{esser2021taming}.
Given an input medical volume $x \in \mathbb{R}^{C \times D \times H \times W}$, an encoder $\mathcal{E}$ maps it to a compressed latent tensor $z_0 = \mathcal{E}(x) \in \mathbb{R}^{C_z \times D' \times H' \times W'}$, which is subsequently reconstructed by a decoder $\mathcal{D}$ as $\hat{x} = \mathcal{D}(z_0)$. The VQ-GAN is trained using a reconstruction objective combined with adversarial and perceptual losses to ensure high-fidelity reconstructions, achieving a compression rate of 8 in each spatial dimension. All diffusion modeling is performed on this latent representation $z_0$.

\begin{figure}[!t]
    \centering
    \includegraphics[width=\textwidth]{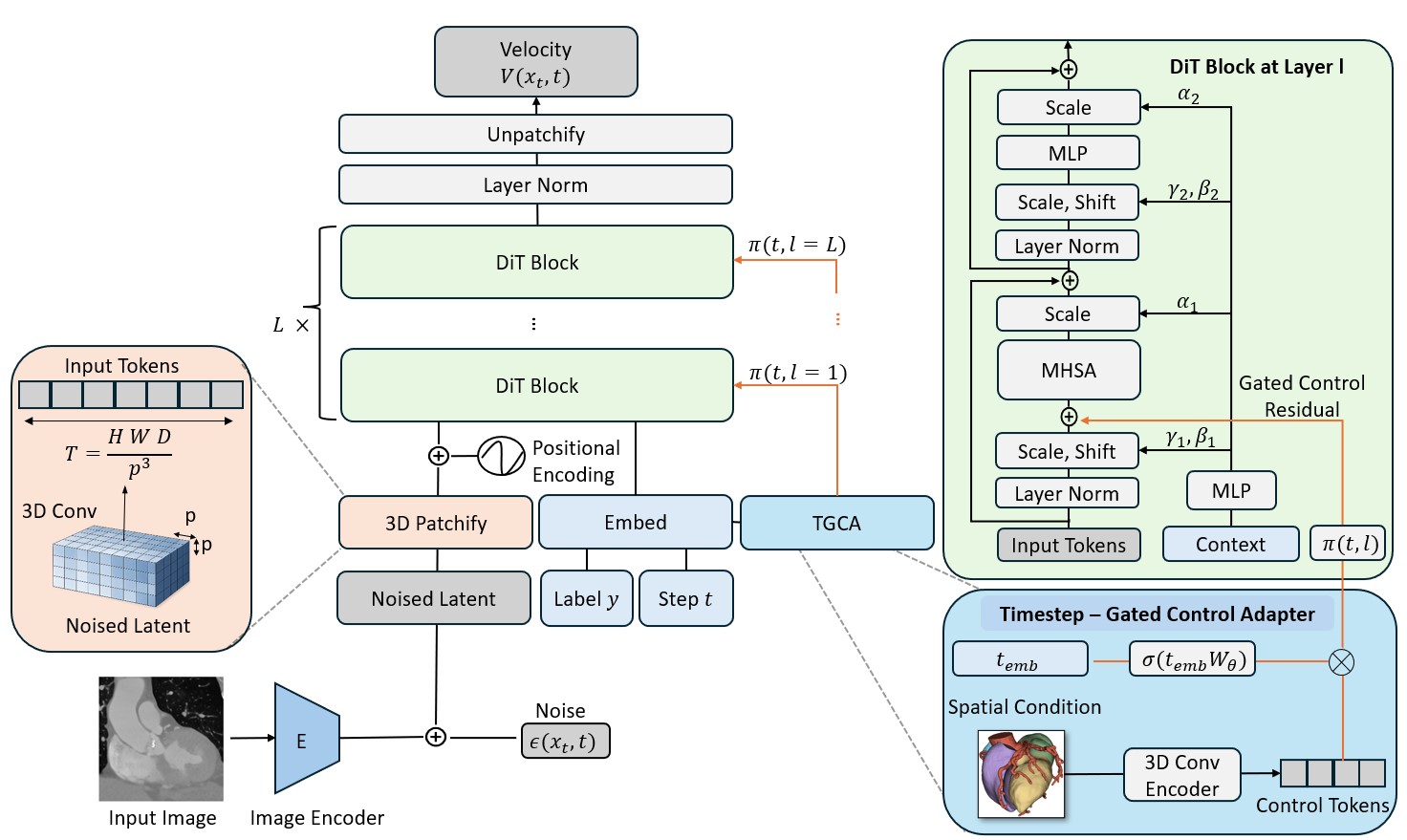}
    \caption{Overview of the VolDiT framework. Input volumes are encoded into latents, patchified into tokens, and processed by the 3D Diffusion Transformer for denoising, with spatial conditioning injected through the timestep-gated control adapter. The transformer outputs are then unpatchified and decoded to reconstruct volumetric images.}
    \label{fig:3D_DiT}
\end{figure}
\subsection{3D Diffusion Transformer}
We train a diffusion model in the latent space of the pretrained VQ-GAN using a cosine noise schedule \cite{nichol2021improved}. The forward process gradually corrupts the latent variables over $T=300$ timesteps. Instead of predicting noise directly, we adopt velocity prediction \cite{salimans2022progressive}. The model is optimized with a Smooth L1 (Huber) loss between the predicted and target velocity. 
To parameterize $v_\theta$, we adapt the design principles of the 2D Diffusion Transformer proposed in computer vision \cite{peebles2023scalable} to volumetric medical imaging. Specifically, we introduce a purely transformer-based 3D denoising backbone operating on volumetric latent tokens.
Given a latent tensor $z_t \in \mathbb{R}^{C_z \times D' \times H' \times W'}$, we partition it into non-overlapping cubic patches of size $p \times p \times p$ (Figure \ref{fig:3D_DiT}). A 3D convolution with kernel size and stride $p$ produces token embeddings $\mathbf{x}_0 \in \mathbb{R}^{T \times d}$, where $T = D'H'W'/p^3$ is the number of tokens and $d$ is the embedding dimension. Fixed 3D sine-cosine positional encodings are added to preserve spatial structure.
The backbone consists of a series of $L$ DiT blocks, following the implementation of \cite{peebles2023scalable}. The final transformer output is projected back to patch-sized latent predictions and reshaped via an unpatchify operation to recover $\hat{v}\theta(z_t, t) \in \mathbb{R}^{C_z \times D' \times H' \times W'}$.
\subsection{Timestep-Gated Control Adapter for Spatial Conditioning}
After unconditional pretraining, we apply spatial conditioning through a lightweight adapter (portrayed in the blue part of Figure \ref{fig:3D_DiT}) while keeping the base DiT backbone frozen \cite{zhang2023adding,t2iadapter}. Given a conditioning volume $s \in \mathbb{R}^{C_s \times D \times H \times W}$ (e.g., a segmentation mask), we encode it with a 3D convolutional network to produce a latent feature map $\mathbf{h}c \in \mathbb{R}^{d \times D' \times H' \times W'}$, explicitly mapping the conditioning input into the DiT token embedding dimension $d$. The feature map is then flattened into a sequence of control tokens $\mathbf{c}{\text{tok}} \in \mathbb{R}^{T \times d}$, where $T = D'H'W'$ is the number of volumetric tokens. The final convolution layer is zero-initialized to ensure stable conditioning training.\\
To adapt the control tokens across diffusion steps, we propose a timestep-dependent gating $\gamma(t)$ computed as the sigmoid of an MLP applied to a time embedding $\mathbf{t}$, i.e., $\gamma(t) = \sigma(\text{MLP}(\mathbf{t}))$. The resulting gating modulates the control tokens as $\tilde{\mathbf{c}}_{\text{tok}} = \gamma(t) \cdot \mathbf{c}_{\text{tok}}$. For selected transformer layers $l$, control is injected as $\mathbf{x}^{(l)} \leftarrow \mathbf{x}^{(l)} + \lambda_l \tilde{\mathbf{c}}_{\text{tok}}$, where $\lambda_l$ is a learnable per-layer scaling parameter. This design enables layer-wise control strength, temporal adaptation across diffusion steps, and stable finetuning without disrupting pretrained weights.
\begin{figure}[!t]
    \centering
    \includegraphics[width=0.7\linewidth]{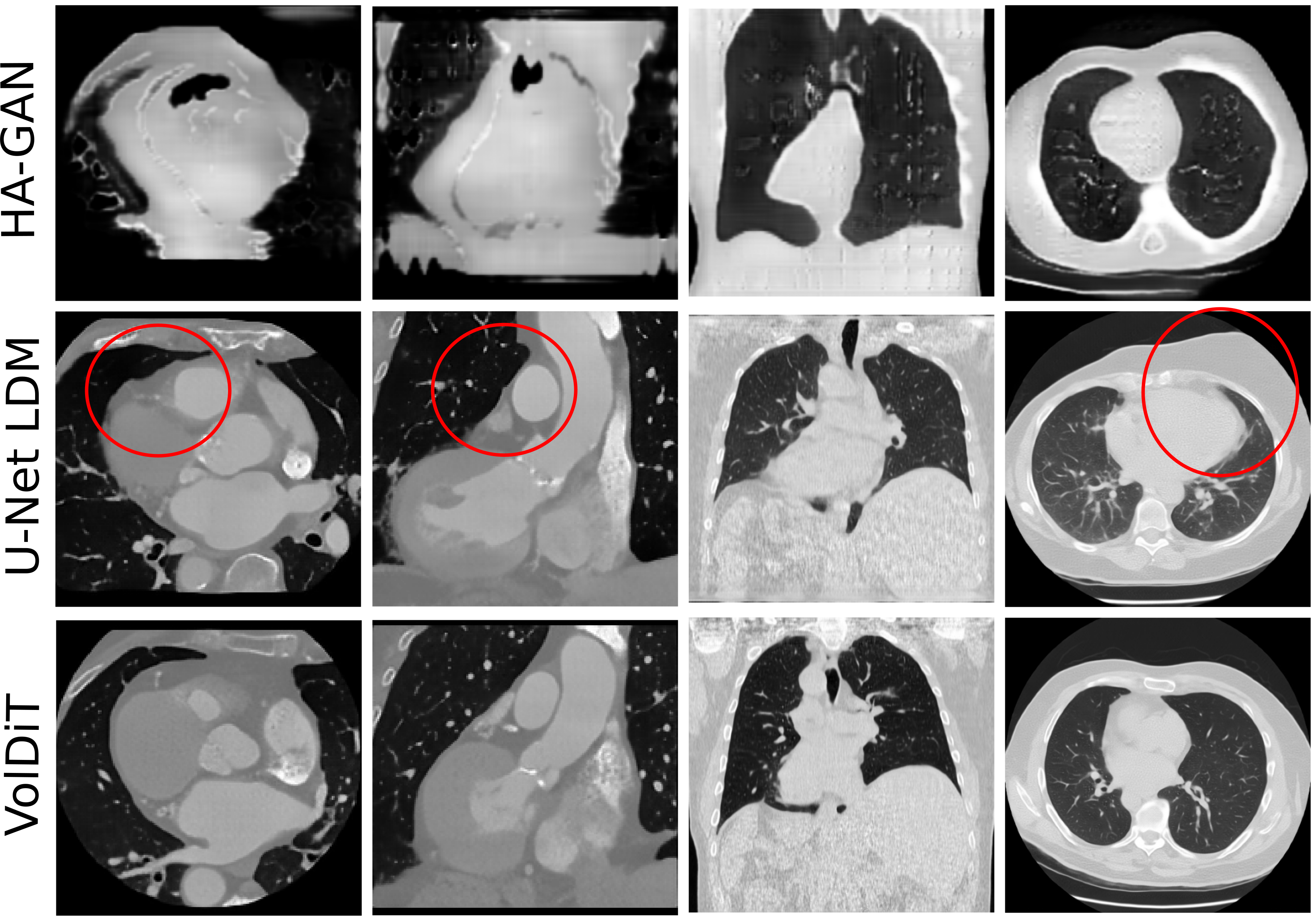}
    \caption{Synthetic examples on LUNA16 and TaviCT. HA-GAN produces anatomically implausible structures, while the U-Net LDM yields high-quality images but struggles with global anatomical consistency (red circles). VolDiT generates anatomically coherent volumes with sharper structural detail, reflecting the benefit of global self-attention.}
    \label{fig:qualitative_uncond}
\end{figure}
\subsection{Datasets}
Our evaluation datasets include:\\
\textbf{Lung CT (LUNA16):} This dataset comprised of 888 lung volumes (3D) from the publicly available Luna16 dataset\footnote{https://luna16.grand-challenge.org/}. Among these, 800 volumes were used for training, 40 volumes for validation, and 48 volumes for testing. The volumes were cropped and padded to $512\times512\times256$, and their intensity clipped to a range of $[-1200, 300]$.\\
\textbf{TaviCT:} This in-house dataset contains 1002 cardiac CTA scans of TAVI patients [anonymized reference], which we split into 600 volumes for training, 201 volumes for validation, and 201 volumes for testing. The volumes were cropped and padded to size $192 \times 192 \times 192$, and their intensity clipped to a range of $[-1000, 1000]$. 

\subsection{Experimental Setup}

Models with minimal validation loss were selected for evaluation. We evaluate generative fidelity using the Fréchet Inception Distance (FID) metric, comparing 100 generated examples against their respective datasets’ test split ($\text{FID}_{\text{test}}$). Additionally, precision, recall, coverage, and density were calculated on the extracted features, where the neighborhood parameter $k$ is selected such that a real--real comparison yields scores greater than 0.95, ensuring a stable and meaningful operating point \cite{prdc} (LUNA16: $k=7$, TaviCT: $k=3$). Due to the absence of publicly available feature extractors specifically trained for cardiac CTA data, we employ ImageNet-pretrained feature extractors for the TaviCT dataset and report a 2.5D FID, computed as the average across the three anatomical axes, which has been shown to yield reliable and stable estimates of generative fidelity in volumetric medical imaging \cite{imagenetvsmednet}. For the LUNA16 dataset, we instead use a 3D ResNet-50 pretrained within MedicalNet on a large corpus of chest CT scans, which aligns well with the lung CT domain of LUNA16 \cite{medicalnet}. Sample diversity is additionally quantified by computing the Multi-Scale SSIM (MS-SSIM) metric across 100 random pairs of the synthetic examples \cite{mssim}.
Lastly, we study the effect of model capacity and patch size on generative quality by evaluating DiT configurations from XS (depth 6, hidden size 384, 6 attention heads) to L, with patch sizes $p=2$ and $p=4$, following \cite{peebles2023scalable}.
We compare VolDiT to a state-of-the-art U-Net-based LDM \cite{seyfarth2025medlord} trained with the same training strategy, i.e., in the same VQ-GAN latent space, and employing the same noise schedule and time steps. In addition, we include HA-GAN as a non-diffusion-based baseline to contrast diffusion-based generative modeling with adversarial training paradigms \cite{hagan}. 
Training procedure and hyperparameters were directly adopted from \url{https://github.com/batmanlab/HA-GAN}.
For conditional synthesis, we use held-out test masks to synthesize guided volumes by utilizing our smallest model, \textit{VolDiT-XS}.
We then evaluate segmentation-conditioned synthesis using Dice similarity, as well as 95th percentile Hausdorff distance between input segmentation masks and segmentations predicted from generated volumes using TotalSegmentator \cite{totalsegmentator}.
To isolate the contribution of timestep-dependent gating, we compare the full model against a variant with constant conditioning strengths $ \pi(t,l)=0.1$ and $\pi(t,l)=1.0$ across diffusion steps.

\begin{table}[!b]
    \centering
    \caption{Generative metrics as measured by FID, precision (P), recall (R), density (D), coverage (C), and MS-SSIM.}
    \begin{tabular}{l|l|cccccccc}
        \hline
        \textbf{Dataset} & \textbf{Model} 
        & \textbf{$\text{FID}_{\text{test}}$}$(\downarrow)$ 
        & \textbf{P}$(\uparrow)$
        & \textbf{R}$(\uparrow)$
        & \textbf{D}$(\uparrow)$ 
        & \textbf{C}$(\uparrow)$ 
        & \textbf{MS-SSIM}$(\downarrow)$ \\
        \hline
        \multirow{4}{*}{LUNA16}
        & Real train  & 0.005 & 0.96 & 0.96 & 0.96 & 1.0 & 0.31 $\pm$ 0.08 \\
        & U-Net LDM \cite{seyfarth2025medlord} & 0.031 & 0.44 & \textbf{0.94} & 0.18 & 0.76 & \textbf{0.36 $\pm$ 0.09} \\
        & HA-GAN \cite{hagan} & 0.126 & 1.00 & 0.00 & 0.32 & 0.06 & 0.99 $\pm$ 0.00 \\
        & \textbf{VolDiT (L, p=4)} & \textbf{0.004} & \textbf{0.91} & 0.90 & \textbf{0.82} & \textbf{0.92} & 0.38 $\pm$ 0.08 \\
        \hline
        \multirow{4}{*}{TaviCT}
        & Real train & 3.1 & 0.95 & 0.95 & 1.07 & 0.90 & 0.30 $\pm$ 0.08 \\
        & U-Net LDM \cite{seyfarth2025medlord} & 36.8 & \textbf{0.95} & 0.63 & 0.89 & 0.68 & 0.38 $\pm$ 0.07 \\
        & HA-GAN \cite{hagan} & 155.5 & 1.00 & 0.00 & 1.25 & 0.04 & 0.99 $\pm$ 0.00 \\
        & \textbf{VolDiT (L, p=2)} & \textbf{21.4} & 0.94 & \textbf{0.73} & \textbf{1.12} & \textbf{0.76} & \textbf{0.35 $\pm$ 0.09} \\
        \hline
    \end{tabular}%
    \label{tab:uncond_qan}
\end{table}
\section{Results}
\subsection{Unconditional Volumetric Synthesis}

We first compare the proposed 3D DiT against the U-Net-based LDM in the unconditional setting. Across both datasets, the results depicted in Table~\ref{tab:uncond_qan} show that VolDiT achieves a more favorable quality–diversity trade-off. On TaviCT, it substantially improves FID (21.4 vs.\ 36.8) while maintaining high precision (0.94 vs.\ 0.95) and increasing recall (0.73 vs.\ 0.63), indicating better mode coverage without sacrificing fidelity. This behavior is further supported by higher density and coverage, suggesting that generated samples both lie closer to the real data manifold and cover a larger portion of it. Additionally, the lower MS-SSIM (0.35 vs.\ 0.38) reflects increased sample diversity. Similar trends are observed on LUNA16, confirming that VolDiT produces synthetic data that is well aligned with real data across datasets.

\subsection{Scaling Behavior and Patch Size Ablation}  
Across our datasets, we do not observe the strictly monotonic improvement with model size reported in 2D computer vision settings \cite{peebles2023scalable}, as shown in Tab. \ref{tab:scaling_compact}. We attribute this to two main factors: (i) the relatively small size of the medical datasets, and (ii) the fixed number of training iterations used for all models. In the original DiT work, larger models consistently outperform smaller ones when trained longer, but in our setting, bigger models may be undertrained and do not always achieve superior FID. Despite these limitations, smaller DiT models already achieve strong unconditional generation performance, supporting the applicability of transformer-based volumetric diffusion for medical imaging.

\begin{table}[!b]
    \centering
    \caption{Effect of model size and patch size (FID$_{\text{test}}$), lower is better)}
    \begin{tabular}{l|l|c|c|c}
        \hline
        \textbf{Dataset} & \textbf{Model} & \textbf{Parameter Count} & \textbf{p=2} & \textbf{p=4} \\
        \hline
        \multirow{4}{*}{LUNA16}
        & VolDiT-XS  & 17.2M  & \textbf{0.009} & 0.015 \\ 
        & VolDiT-S   & 33.2M  & \textbf{0.009} & 0.023 \\
        & VolDiT-B   & 131.0M & 0.014 & 0.017 \\
        & VolDiT-L   & 580.0M & --  & \textbf{0.004} \\
        \hline
        \multirow{4}{*}{TaviCT}
        & VolDiT-XS  & 17.2M  & 22.0 & 134.6 \\
        & VolDiT-S   & 33.2M  & 23.6 & 131.2\\
        & VolDiT-B   & 131.0M & 24.2 & 125.8\\
        & VolDiT-L   & 580.0M & \textbf{21.4} & \textbf{33.3}\\
        \hline
    \end{tabular}%
    \label{tab:scaling_compact}
\end{table}

\subsection{Segmentation-Conditioned Synthesis on TaviCT}
As shown in Table~\ref{tab:conditional_tavi_scales}, VolDiT with gated control significantly improves anatomical alignment compared to the U-Net-based latent diffusion baseline, without sacrificing generative fidelity. While selecting a fixed conditioning scale requires careful manual tuning, e.g., ($\pi_{0.1}$) leads to weaker anatomical alignment. On the contrary, a high fixed scale ($\pi_{1.0}$) improves Dice and HD${95}$, but risks degrading fidelity due to over-constraining the generative prior. The learnable gating strategy ($\pi\theta$) allows the model to adaptively exploit the full conditioning capacity while maintaining the best FID and competitive diversity. This behavior indicates balanced residual modulation rather than structural override and confirms that adaptive temporal scaling stabilizes conditional diffusion in volumetric token space.

\begin{table}[!b]
    \centering
    \caption{Conditional generation performance on TaviCT using held-out test masks. Median (IQR) reported for Dice and HD$_{95}$. Lower FID and MS-SSIM indicate better fidelity and higher diversity, respectively.}
    \begin{tabular}{l|cc|cc|cc}
        \hline
        & \multicolumn{2}{c|}{Heart} 
        & \multicolumn{2}{c|}{Aorta} 
        & \multicolumn{2}{c}{Gen. Metrics} \\
        \cline{2-7}
        \textbf{Model} 
        & Dice $\uparrow$ 
        & HD$_{95}$ $\downarrow$ 
        & Dice $\uparrow$ 
        & HD$_{95}$ $\downarrow$
        & FID$_{\text{test}}$ $\downarrow$
        & MS-SSIM $\downarrow$ \\
        \hline
        U-Net LDM 
            & 0.89 (0.03) 
            & 5.39 (2.0)
            & 0.83 (0.10) 
            & 4.69 (2.0)
            & 25.8
            & 0.34 $\pm$ 0.09 \\
        VolDiT ($\pi_{0.1}$) 
            & 0.92 (0.04)
            & 4.12 (6.7)
            & 0.90 (0.10)
            & 2.83 (4.1)
            & 25.1
            & 0.36 $\pm$ 0.07 \\
        VolDiT ($\pi_{1.0}$) 
            & \textbf{0.94 (0.02)}
            & \textbf{2.83 (0.8)}
            & \textbf{0.92 (0.02)}
            & 2.00 (0.5)
            & 24.4
            & \textbf{0.33 $\pm$ 0.08} \\
        VolDiT ($\pi_\theta$) 
            & \textbf{0.94 (0.02)}
            & 2.83 (0.9)
            & \textbf{0.92 (0.02)}
            & \textbf{2.00 (0.3)}
            & \textbf{22.7}
            & 0.34 $\pm$ 0.08 \\
        \hline
    \end{tabular}%

    \label{tab:conditional_tavi_scales}
\end{table}

\begin{figure}[!t]
    \centering
    \includegraphics[width=0.9\textwidth]{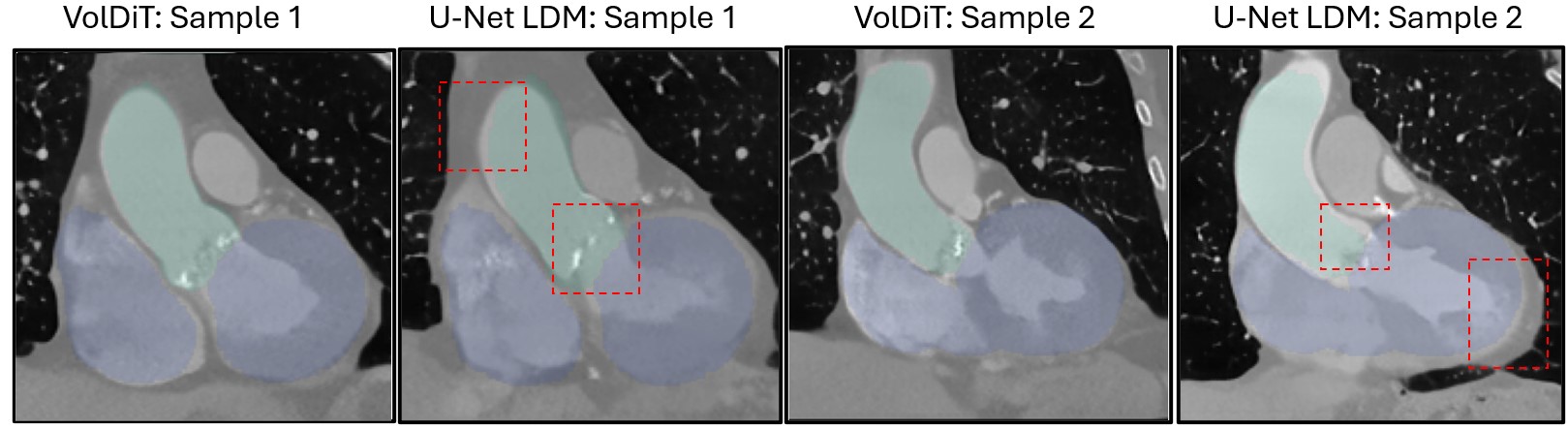}
    \caption{Conditionally generated TaviCT samples based on held-out test masks of heart structures (blue) and the aorta (green). The U-Net-based model shows less alignment with input masks and does not preserve anatomical realism when enforcing the condition (highlighted by red squares). }
    \label{fig:placeholder}
\end{figure}

%
%

%
%
%

\section{Conclusion}
In this work, we investigated whether a fully transformer-based diffusion backbone can effectively model native 3D volumetric medical data. By directly learning the full 3D latent distribution using global self-attention over volumetric tokens, VolDiT achieves anatomically coherent synthesis within a conceptually simple architecture. Quantitative and qualitative results demonstrate improved fidelity and diversity compared to convolutional U-Net-based latent diffusion baselines.
A key design principle of our approach is architectural unification: we rely on a purely token-based backbone operating on the entire 3D latent volume. In this sense, VolDiT provides empirical evidence that diffusion transformers extend naturally beyond 2D image synthesis to higher-dimensional medical data.
Although this study focused primarily on 3D volumes, the framework establishes the foundation for high-dimensional generative modeling, as demonstrated in our concurrent extension (see Supp. 1). Additionally, while our evaluations are restricted to a single autoencoder architecture, the approach is not limited to this choice. VolDiT can, in principle, be combined with alternative latent representations, such as autoencoders trained as foundation models \cite{guo2025maisi} or wavelet-based transformations \cite{friedrich2024wdm}, potentially further improving efficiency and generalization.
Furthermore, we applied a timestep-gated control adapter \textit{TGCA} for segmentation-conditioned synthesis. By mapping spatial conditions into token space and learning weighted residual modulations of transformer features, the adapter enables stable and anatomically precise control without modifying the pretrained backbone. Importantly, this design is modality-agnostic: any spatial prior encoded by a pretrained model can, in principle, be projected into token space and integrated via TGCA, facilitating the incorporation of foundation-model linking or multimodal conditioning signals. Finally, by establishing a robust, fully transformer-based baseline for 3D medical data, VolDiT provides the necessary framework for future research on medical image synthesis, such as comparing memorization effects of transformer-based architectures against those observed in traditional U-Net-based LDMs~\cite{Dar2025}.
\subsubsection{\ackname}
This work was supported by Heidelberg Faculty of Medicine at Heidelberg University, by the Multi-DimensionAI project of the Carl Zeiss Foundation (P2022-08-010) and by the EU-Horizon Project \emph{DVPS} (101213369). The authors acknowledge \emph{1-} the data storage service SDS@hd supported by the Ministry of Science, Research and the Arts Baden-Württemberg (MWK) and the German Research Foundation (DFG) through grant INST 35/1314-1 FUGG and INST 35/1503-1 FUGG, \emph{2-} the state of Baden-Württemberg through bwHPC and the DFG through grant INST 35/1597-1 FUGG.  

%
%
%
\bibliographystyle{splncs04}
\bibliography{3D_DiT.bib}

%
%
%
%
%

\end{document}